\pdfoutput=1

\documentclass[11pt]{article}

\usepackage[preprint]{acl}

\usepackage{times}
\usepackage{latexsym}

\usepackage[T1]{fontenc}

\usepackage[utf8]{inputenc}

\usepackage{microtype}

\usepackage{inconsolata}

\usepackage{graphicx}

\usepackage{algorithm}
\usepackage{algpseudocode}
\usepackage{amsmath}
\usepackage{booktabs}
\usepackage{multirow}
\usepackage{subcaption}

\newcommand{\renyi}[0]{R\'enyi\ }
\DeclareMathOperator{\ent}{H}

\title{How Well Does First-Token Entropy Approximate Word Entropy as a Psycholinguistic Predictor?}

\author{Christian Clark \\ The Ohio State University \\ \texttt{clark.3664@osu.edu}
        \And  Byung-Doh Oh \\ New York University \\ \texttt{oh.b@nyu.edu}
        \And William Schuler \\ The Ohio State University \\ \texttt{schuler.77@osu.edu}}

\begin{document}
\maketitle
\begin{abstract}
Contextual entropy is a psycholinguistic measure capturing the anticipated difficulty of processing a word just before it is encountered.
Recent studies have tested for entropy-related effects as a potential complement to well-known effects from surprisal.
For convenience, entropy is typically estimated based on a language model's probability distribution over a word's first subword token.
However, this approximation results in underestimation and potential distortion of true word entropy.
To address this, we generate Monte Carlo (MC) estimates of word entropy that allow words to span a variable number of tokens.
Regression experiments on reading times show divergent results between first-token and MC word entropy, suggesting a need for caution in using first-token approximations of contextual entropy.

\end{abstract}

\section{Introduction}

Recent studies of human sentence processing have explored potential psycholinguistic effects from the contextual entropy of the current word being processed \cite{cevolietal22,pimenteletal23,wilcoxetal23,giulianellietal24}.
Contextual entropy is an information-theoretic measure quantifying a reader's level of uncertainty about the current word based on the preceding context; it is computed purely based on the conditional probability distribution of the current word without reference to the word's identity.
As such, it contrasts with surprisal, another commonly used psycholinguistic measure whose effects can be understood as integration costs for an already observed word \cite{cevolietal22,pimenteletal23}. 

Contextual entropy is commonly estimated using a language model (LM) like GPT2 \cite{radfordetal19}.
However, because words can span multiple subword tokens in an LM's vocabulary---making a full summation over their probability distribution intractable---entropy is typically calculated over the probability distribution of each word's first token instead.
This practice results in a systematic underprediction of true word entropy \cite{pimenteletal23}, which is magnified in contexts in which multi-token words are probable.
This mismatch between the phenomenon of interest (word-level predictive processing) and the metric of choice (token-level entropy) may lead to challenges in drawing psycholinguistic conclusions from experimental results.

To address this issue, we calculate LM-based entropy estimates using a technique based on Monte Carlo \citep[MC;][]{metropolisulam49} sampling that allows words to span multiple tokens.
While this method still only provides an approximation of true word entropy, the MC method produces unbiased estimates \citep{giulianellietal24}---unlike first-token approximation---and it can produce higher-quality (lower-variance) estimates if the number of samples is increased.

To evaluate the difference between first-token entropy and MC word entropy, both measures are evaluated in a set of regression experiments using naturalistic self-paced reading and eye-tracking data.
Results show contrasting predictions from the two measures, with the MC word entropy especially making stronger estimates of self-paced reading times.
The gap between the two entropy estimates suggests that first-token entropy may not provide a reliable approximation of true word entropy for psycholinguistic modeling.\footnote{Code for first-token and Monte Carlo entropy estimation is available at \url{https://github.com/christian-clark/word-entropy}.}

\section{Related Work}

Much of the interest in information-theoretic predictors of human sentence processing traces back to surprisal theory \citep{hale01,levy08}, which posits a direct link between processing difficulty and a word's surprisal (negative log probability).
Robust effects from surprisal have been observed across multiple languages and psycholinguistic measures \citep[e.g.,][]{wilcoxetal23,shainetal24}.
Entropy-based predictors are thus often evaluated as possible complements to well-established surprisal predictors; we follow this practice by including surprisal predictors in our regression models.

Several forms of entropy have been evaluated in earlier work.
Word-level (or token-level) contextual entropy, the predictor of focus in the present study, is studied by \citet{vanschijndellinzen19} as well as several more recent studies \citep{cevolietal22,pimenteletal23,wilcoxetal23,giulianellietal24}.
Other work considers the total entropy of the remainder of a sentence---its raw value \citep{roarketal09}, the reduction in this entropy at each incoming word \citep{hale03,hale06}, or both \citep{linzenjaeger16}.
Because total entropy is difficult to exactly compute---especially with contemporary language models---some additional work calculates the entropy of a bounded number of future words as a middle ground \citep{frank13,giulianellietal24}.

The work by \citet{giulianellietal24} comes
closest to our proposed method of MC estimation of contextual word entropy.
These authors approximate the entropy of a continuation of a sentence by generating samples of up to $L\in \{ 5, 10, 15\}$ tokens following a context string.
This differs from our work in that they aim to approximate continuation entropy rather than next-word entropy.

\section{Formulations of Contextual Entropy}

\subsection{Shannon Entropy}
Shannon entropy \citep{shannon48} is the standard form of entropy studied in previous psycholinguistic work. 
A word's contextual Shannon entropy is defined as its expected surprisal, i.e., its expected negative log probability given the preceding words $w_{1..i-1}$: 
\begin{align}
    &\ent(W_i \mid w_{1..i-1}) = \\ \nonumber
    &-\sum_{w \in V} P(w \mid w_{1..i-1}) \log_2 P(w \mid w_{1..i-1}),  
\end{align}
where $V$ is the vocabulary of all possible words and $W_i$ is a random variable over $V$.

\subsection{R\'enyi Entropy}
\citet{pimenteletal23} discuss the possibility that readers' anticipatory processing may be guided by strategies other than considering the expected surprisal of the next word (i.e., Shannon entropy).
For instance, readers might rely on the surprisal of the single most likely word; or, at the other extreme, they might consider the number of possible next words regardless of each word's exact probability.

\renyi entropy \citep{renyi61} is a generalization of Shannon entropy which captures this spectrum of possible anticipatory reading strategies.
A word's contextual \renyi entropy of order $\alpha$ is defined as follows: 
\begin{align} \label{eq:renyi}
    \ent_\alpha (&W_i \mid w_{1..i-1}) = \\ \nonumber
    &\lim_{\beta \rightarrow \alpha} \frac{1}{1-\beta} \log_2 \sum_{w \in V} \Big( P(w \mid w_{1..i-1}) \Big)^\beta.  
\end{align}

When $\alpha=0$, \renyi entropy considers the number of possible next words.
If $\alpha=1$, \renyi entropy equals Shannon entropy (via an application of L'H\^opital's rule to Eq.~\eqref{eq:renyi}).
And when $\alpha=\infty$, \renyi entropy measures the surprisal of the single most likely word in context.

\citet{pimenteletal23} test several values of $\alpha$ and find that \renyi contextual entropy with $\alpha=1/2$ is a relatively strong predictor of reading times.
Our regression experiments evaluate both Shannon entropy and \renyi entropy with $\alpha=1/2$ to get a fuller picture of how first-token and MC approximation affect metrics describing anticipatory word processing.

\subsection{First-Token and MC Approximations}

Contemporary LMs typically work with a finite subword vocabulary---defined using a method like Byte-Pair Encoding \citep{sennrichetal16}---that supports an infinite word vocabulary, as words can span a variable number of subwords.
Because summing over an infinite vocabulary is intractable, recent work instead considers the entropy of a word's first subword token.
First-token contextual Shannon entropy is defined as follows:
\begin{align}
    &\ent(W_i \mid w_{1..i-1}) \approx \\ \nonumber
    &-\sum_{t \in T} P(t \mid w_{1..i-1}) \log_2 P(t \mid w_{1..i-1}),  
\end{align}
where $t$ is a token drawn from the LM's subword vocabulary $T$.
First-token \renyi entropy can be defined analogously.
These approximation are a lower bound on true word entropy \citep{pimenteletal23}.

MC estimation of contextual word entropy is performed by computing the average surprisal of a set of words $S$ sampled in context $w_{1..i-1}$:
\begin{align}
    &\ent(W_i \mid w_{1..i-1}) \approx -\frac{1}{\lvert S \rvert} \sum_{s \in S} \log_2 P(s \mid w_{1..i-1}).  
\end{align}
Each sampled word $s \in S$ is produced by randomly generating successive subword tokens until a word boundary (i.e., the subsequent whitespace in English) is reached.
The surprisal of $s$ is calculated by summing over the surprisal of its subword tokens.

An MC estimate of \renyi entropy can be obtained by replacing the summation term in Equation~\ref{eq:renyi} with a sample-based approximation:
\begin{align}
    &\ent_\alpha (W_i \mid w_{1..i-1}) \approx \\ \nonumber
    &\lim_{\beta \rightarrow \alpha} \frac{1}{1-\beta} \log_2 \Bigg(\frac{1}{\lvert S \rvert} \sum_{s \in S} \Big( P(s \mid w_{1..i-1}) \Big)^{\beta-1} \Bigg).  
\end{align}

Due to limitations in the available compute budget, all MC estimates in this work restrict the number of samples to $\lvert S \rvert = 512$.
However, an analysis of sample variance indicated that entropy estimates are reasonably stable with this sample count.
See Appendix \ref{appx:mc} for this analysis and other details about the MC estimation.

\begin{figure}[t]
    \centering
    \includegraphics[width=\linewidth]{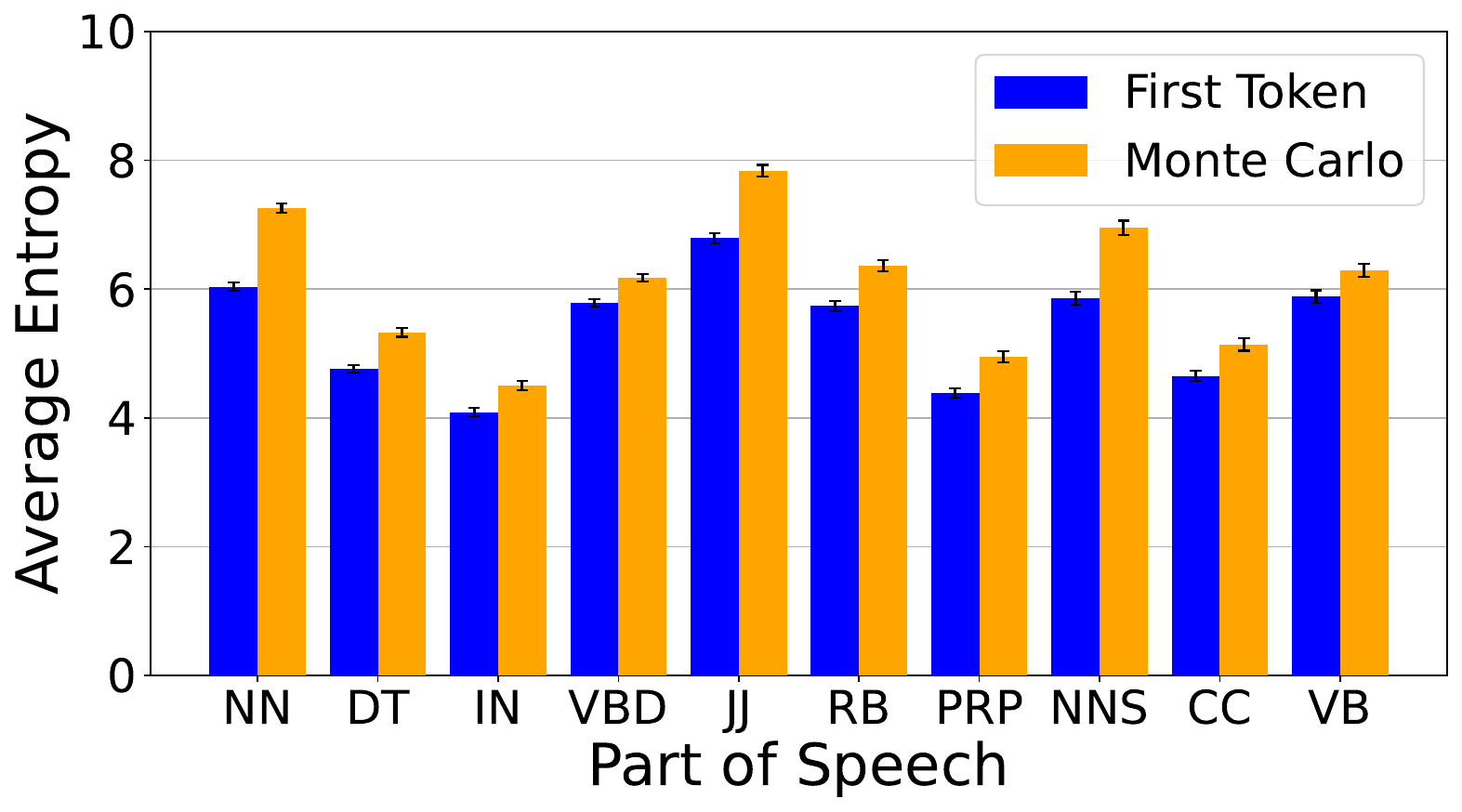}
    \caption{Average Shannon entropy of the 10 most frequent parts of speech in the Natural Stories corpus, using either first-token or Monte Carlo word entropy approximations. Part-of-speech tags are from Penn Treebank annotations \cite{marcusetal93}. Error bars represent \textpm1 standard error of the mean (SEM).}
    \label{fig:posentropy}
\end{figure}

To illustrate the difference between first-token and MC word entropy, Figure~\ref{fig:posentropy} shows the average Shannon entropies of the 10 most frequent parts of speech in the Natural Stories corpus \citep{futrelletal21}, one of the English psycholinguistic corpora used in subsequent regression experiments  (Sec.~\ref{sec:regression}).
As expected, MC word entropy is consistently higher than first-token entropy.
It can also be observed that open-class parts of speech such as NN (noun) and JJ (adjective) show a larger relative difference between first-token and MC word entropy compared to closed classes like IN (preposition).
This likely reflects a wider range of multi-token words available within these part-of-speech categories; it also provides evidence that the first-token approximation not only underpredicts but also distorts true word entropy.

\begin{table}[t!]
    \centering
    \small
    \begin{tabular}{lrr} \toprule
    Corpus & Fit & Held-out \\ \midrule
    Natural Stories SPR & 384,984 & 192,826 \\
    Brown SPR & 59,292 & 29,671 \\
    \midrule
    Dundee FP & 98,115 & 48,598 \\
    Dundee GP & 98,115 & 48,598 \\
    Provo FP & 52,959 & 26,539 \\
    Provo GP & 52,960 & 26,539 \\
    GECO FP & 144,850 & 72,468 \\
    GECO GP & 144,850 & 72,468 \\
    \bottomrule
    \end{tabular}
    \caption{Number of observations in the fit and held-out partitions of each reading-time corpus.}
    \label{tab:observations}
\end{table}

\section{Regression Experiments} \label{sec:regression}

To compare the psycholinguistic predictive power of first-token entropy and word entropy, we perform linear mixed-effects \citep[LME;][]{batesetal15} regression experiments on a set of naturalistic reading-time corpora in English.

\subsection{Corpora}

The psycholinguistic corpora included two self-paced reading (SPR) corpora and three corpora with first-pass (FP) and go-past (GP) durations from eye tracking.
The self-paced reading corpora were Natural Stories \citep{futrelletal21}, which contains data from 181 subjects who read 10 naturalistic stories; and Brown \citep{smithlevy13}, which contains data from 35 subjects who read 13 passages from the Brown corpus.
The eye-tracking corpora were Dundee \citep{kennedyetal03}, containing fixation durations from 10 subjects who read 67 newspaper editorials; Provo \citep{lukechristianson18}, containing fixation durations from 84 subjects who read 55 passages from a variety of sources including news articles, magazines, and works of fiction; and GECO \citep{copetal17}, containing fixation durations from 14 subjects who read a 13-chapter Agatha Christie novel \citep{christie20}.

Each corpus was divided into fit and held-out partitions.
The fit partition was used for fitting LME regression models, and the held-out partition was used to compute log likelihood scores.
Table~\ref{tab:observations} shows the number of observations per partition in each corpus.

\subsection{Predictors}

Baseline predictors in the LME models included word length, word index, unigram surprisal, LM surprisal of the current and previous word (SPR, FP, and GP), and whether the previous word was fixated (FP and GP only).
Unigram surprisal was estimated on the 6.5 billion whitespace-delimited words from the OpenWebText Corpus \citep{gokaslancohen19} using the KenLM toolkit \citep{heafieldetal13}.
SPR regression models included per-subject random slopes for word length, word index, and LM surprisal (current and previous word), and a per-subject random intercept.
Regression models for eye-tracking only included a per-subject random intercept.
Other random slopes were removed to ensure convergence.

For each corpus and response type, the difference in log likelihood ($\Delta$LL) was calculated between a regression model containing only the baseline predictors, and a regression model additionally containing an entropy predictor (either first-token entropy or MC-based word entropy).
This evaluation was conducted twice, once using Shannon entropy and once using \renyi entropy ($\alpha=1/2)$.
GPT2-small was the LM used to calculate entropy and surprisal predictors.

\begin{table}[t]
        \centering
        \footnotesize
        \begin{tabular}{lrrrr}
            \toprule
            & \multicolumn{2}{c}{Shannon entropy} & \multicolumn{2}{c}{R\'enyi entropy} \\
            Corpus & $\Delta$LL$_{\text{FT}}$ & $\Delta$LL$_{\text{MC}}$ & $\Delta$LL$_{\text{FT}}$ & $\Delta$LL$_{\text{MC}}$ \\
            \midrule
            NS SPR & 29 & 72 & 60 & 272 \\
            Brown SPR & 1.4 & 7.9 & 1.5 & 13  \\
            \midrule
            Dundee FP & 1.3 & 0.6 & 0.3 & 2.2 \\
            Dundee GP & 1.2 & 0.0 & 0.2 & 9.3 \\
            Provo FP & $-$0.2 & 0.7 & 0.0 & $-$0.5  \\
            Provo GP & 1.6 & 1.1 & $-$0.4 & 2.3 \\
            GECO FP & 3.6 & $-$0.4 & $-$0.4  & 3.1 \\
            GECO GP & $-$0.2 & $-$0.4 & 0.0 & 1.3 \\
            \bottomrule
        \end{tabular}
    \caption{Increases in log likelihood from adding a target entropy predictor to a baseline regression model for predicting self-paced reading (SPR) time, first-pass (FP) duration, or go-past (GP) duration. $\Delta$LL$_\text{FT}$ and $\Delta$LL$_\text{MC}$ respectively refer to log likelihood improvements from first-token and MC entropy approximations. NS means Natural Stories.}
    \label{tab:loglik}
\end{table}

\subsection{Results}

Table~\ref{tab:loglik} presents the results from this experiment.
When using Shannon entropy, replacing first-token estimates with MC estimates improves $\Delta$LL scores in the two self-paced reading corpora.
However, the opposite pattern is observed on most eye-tracking corpora.
In some cases, $\Delta$LL is negative, indicating that including an entropy predictor hurt eye-tracking predictions.
The generally small $\Delta$LL values on eye tracking suggest that word entropy may be a stronger predictor of self-paced reading times than eye-tracking measures.
The muted results for eye tracking may also reflect our choice of relatively strong baseline predictors (e.g., LM surprisal of the current and previous word).

Compared to first-token estimation, MC estimation of \renyi entropy improves predictions on the two SPR corpora, and on 5 out of 6 eye-tracking evaluations.
This seems to suggest that---unlike Shannon entropy---word-level estimates of \renyi entropy are consistently better predictors of anticipatory processing.

To evaluate whether the observed differences between first-token and MC entropy were significant, permutation tests were run over squared errors aggregated over all reading time corpora.
Separate significance tests were run for the Shannon and \renyi entropy predictors.
Both entropy variants showed significant differences between the first-token and MC predictors (Shannon entropy: $p<0.01$; \renyi entropy: $p<0.001$), suggesting that word entropy is more strongly correlated with reading behavior than first-token entropy.

\section{Discussion}
First-token approximations of contextual entropy have often been used as a tool to study anticipatory word-level processing \citep{cevolietal22,pimenteletal23,wilcoxetal23}.
However, this work shows that unbiased estimates of word entropy from MC sampling often lead to experimental results that diverge from those using first-token entropy; in many cases, MC estimates provide a closer match to human behavioral data. 
The concrete difference across the two conditions warrants caution against using first-token entropy in psycholinguistic modeling.

\section*{Limitations}

Concerns about first-token entropy estimates are relevant for contemporary Transformer LMs that use a vocabulary of subword tokens; however, they do not apply to word-level LMs, such as the LSTM used by \citet{vanschijndellinzen19} to study contextual entropy.

Even in subword-based models, the degree to which first-token entropy distorts true word entropy may be tied to the size of an LM's subword vocabulary; LMs with a larger subword vocabulary may have a closer to 1:1 relationship between tokens and words and thus show less distortion.

\bibliography{bibliography}

\appendix

\begin{figure}[t!]
    \centering
    \begin{subfigure}[b]{0.4\textwidth}
        \centering
        \includegraphics[width=\linewidth]{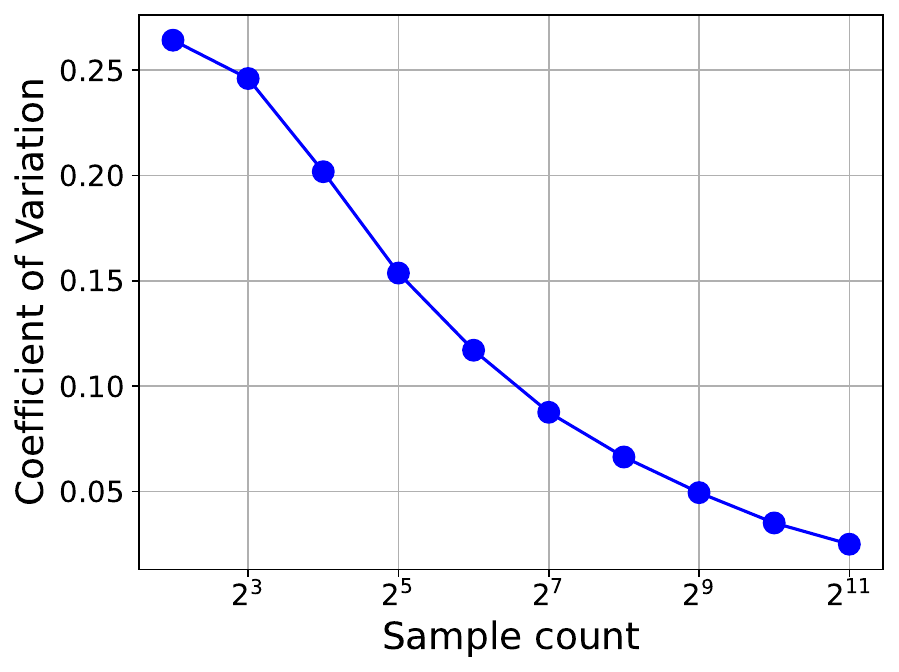}
        \caption{Shannon entropy}
        \hfill
    \end{subfigure}
    \begin{subfigure}[b]{0.4\textwidth}
        \centering
        \includegraphics[width=\linewidth]{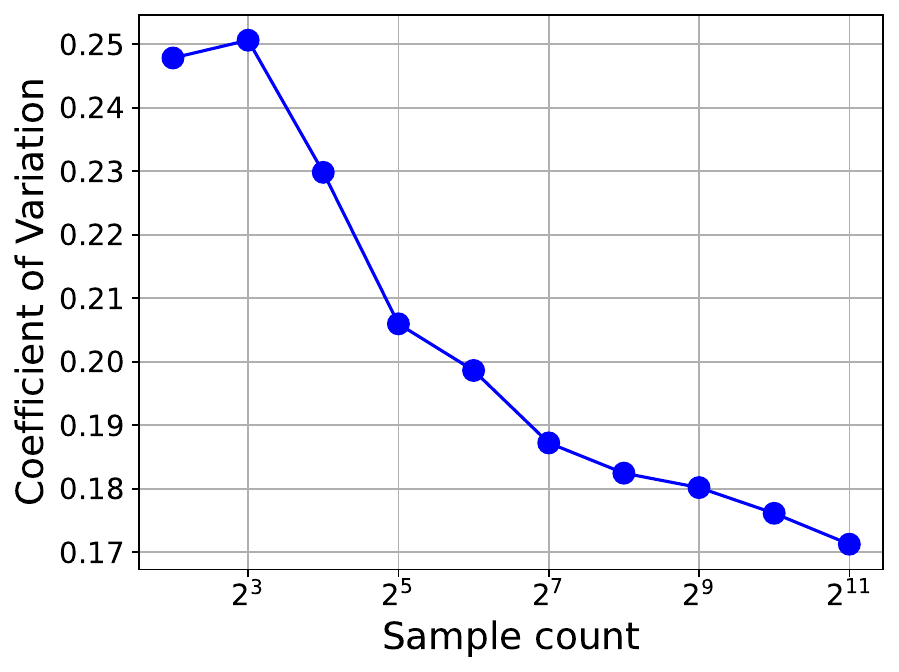}
        \caption{\renyi entropy $(\alpha=1/2)$}
    \end{subfigure}
    \caption{Coefficient of variation by sample count for Monte Carlo estimates of word entropy.}
    \label{fig:cv}
\end{figure}

\section{Monte Carlo Sampling Details} \label{appx:mc}

\subsection{Sampling Procedure} \label{appx:sampling}

Sampling a potentially multi-token word $w_i$ from a language model's conditional probability distribution $P(W_i \mid w_{1..i-1})$ involves iteratively sampling the subword tokens of $w_i$ until a word boundary is reached.
A challenge that arises is that LMs like GPT2 treat word boundaries as leading whitespaces on tokens, meaning that the end-of-word boundary for $w_i$ will be part of the first token for $w_{i+1}$.
These word boundaries must be carefully managed in order to ensure that probabilities of multi-token words form a proper distribution \citep{ohschuler24,pimentelmeister24}. 

To properly track word boundaries, we follow \citet{ohschuler24} in separating the subword vocabulary $T$ of an LM into  a subset containing whitespace-initial tokens $T_B$ and a subset containing tokens with no initial whitespace $T_I$.
Note that $T = T_B \cup T_I$ and $T_B \cap T_I = \emptyset$.
To sample $w_i$, we first sample a word-initial token from $T_B$, with LM probabilities renormalized to sum to 1 over $T_B$.
Subsequent tokens are drawn from $T_I \cup \{\textsc{eow}\}$, where tokens in $T_I$ are assigned their usual conditional probabilities, and 
$$P(\textsc{eow} \mid w_{i..i-1}, t_{1..j}) = \sum_{t \in T_B} P(t\mid w_{i..i-1}, t_{1..j}),$$
where $t_{1..j}$ are the subword tokens that have been sampled so far.
In other words, the probability of \textsc{eow} is the probability of sampling any whitespace-initial token, which means the end of $w_i$ has been reached.
The surprisal of $w_i$ is the sum of the surprisal of the initial token from $T_B$, any middle tokens from $T_I$, and the final \textsc{eow}.

To ensure that sampling was tractable in our experiments, the number of possible subword tokens in a word was capped at 20.
Strictly speaking, this means that the MC estimates in this work will tend to underestimate true surprisal.
However, words with more than 20 tokens are exceptionally unlikely to sample---empirically, we observe that around 1/50,000 samples reach 20 tokens---so the effect on MC estimates should be minimal. 

\subsection{Variance Analysis}

To measure the effect of sample count on the variance of the Monte Carlo estimates, we performed a bootstrapping \citep{efron92} analysis using the first story from the Natural Stories corpus.
We followed a similar procedure to \citet[Sec.~5.1]{giulianellietal24}.
First, for each $k \in \{2^j \mid j = 2, 3, \dots, 11\}$ and word position $i$ in the story, a set of $k$ word samples $S_{k,i}$ was taken following the procedure in \ref{appx:sampling}.
Next, a set of 1000 resamples with replacement---each of size $k$---were taken from each $S_{k,i}$, and entropy was calculated over each resample.
The coefficient of variation for word $i$ was then calculated as $CV_{k,i} = \sigma_{k,i}/\mu_{k,i}$, where $\sigma_{k,i}$ and $\mu_{k,i}$ are respectively the standard deviation and mean of the entropy across all resamples.
Finally, the average coefficient of variation with $k$ samples, $CV_k$, was found by averaging over all $CV_{k,i}$ values.

Figure~\ref{fig:cv} plots the $CV_k$ value for each sample size $k$ for both Shannon and \renyi entropy.
Generally speaking, \renyi entropy requires more samples than Shannon entropy to attain a given coefficient of variation.
Both entropy measures benefit from higher sample counts, but the $CV_k$ values are reasonably stable near the sample count of $2^9 = 512$ used in the regression experiments (Sec.~\ref{sec:regression}).

\end{document}